\documentclass[journal,twoside,web]{ieeecolor}
\usepackage{generic}
\usepackage{cite}
\usepackage{amsmath,amssymb,amsfonts}
\usepackage{graphicx}
\usepackage{hyperref}
\hypersetup{hidelinks=true}
\usepackage{textcomp}

\usepackage{amsmath, amssymb, bm}

\usepackage{booktabs,tabularx}
\usepackage{makecell}
\usepackage{threeparttable}

\usepackage{array,multirow}
\newcolumntype{L}[1]{>{\raggedright\arraybackslash}p{#1}}
\usepackage{multicol}

\usepackage{xcolor}
\newcommand{\rev}[1]{\textcolor{black}{#1}}

\def\BibTeX{{\rm B\kern-.05em{\sc i\kern-.025em b}\kern-.08em
    T\kern-.1667em\lower.7ex\hbox{E}\kern-.125emX}}

\begin{document}

\setlength{\textfloatsep}{8pt plus 2pt minus 2pt}
\setlength{\dbltextfloatsep}{7pt plus 2pt minus 2pt}
\setlength{\floatsep}{5pt plus 2pt minus 2pt}
\setlength{\dblfloatsep}{5pt plus 2pt minus 2pt}
\setlength{\intextsep}{5pt plus 2pt minus 2pt}
\setlength{\abovecaptionskip}{3pt}
\setlength{\belowcaptionskip}{0pt}
\renewcommand{\textfraction}{0.05}
\renewcommand{\topfraction}{0.95}
\renewcommand{\bottomfraction}{0.95}
\renewcommand{\floatpagefraction}{0.85}
\renewcommand{\dbltopfraction}{0.95}
\renewcommand{\dblfloatpagefraction}{0.85}
\setcounter{topnumber}{3}
\setcounter{dbltopnumber}{3}
\setcounter{bottomnumber}{2}
\setcounter{totalnumber}{5}

\title{Time Window Optimization for Motor Imagery Classification via Temporal Structure Embedded Reinforcement Learning}
\author{Huan Luo, Na Lu, Xiaopeng Wang, and Ziyang Shang
\thanks{This work was supported by National Natural Science Foundation of China under Grant No. 62476213, Natural Science Basic Research Program of Shaanxi Province under Grant 2024JC-YBMS-486. \textit{(Corresponding author: Na Lu.)}}
\thanks{Huan Luo is with the School of Future Technology, Xi’an Jiaotong University, Xi'an, 710049, China (e-mail: luohuan123@stu.xjtu.edu.cn). }
\thanks{Other authors are with the School of Automation Science and Engineering, Xi’an Jiaotong University, Xi'an, 710049, China (e-mail: lvna2009@mail.xjtu.edu.cn; wxp035112@stu.xjtu.edu.cn; 2203712691@stu.xjtu.edu.cn).}
}

\maketitle

\begin{abstract}
Online motor imagery-based brain-computer interfaces (MI-BCIs) usually processes continuous EEG signal with fixed-length sliding windows, which is inflexible, human intervened, and misaligned with neural events. Time window optimization has shown promising improvement as a substitute solution. However, the existing optimization methods typically evaluate windows isolatedly and fail to model temporal dependencies across consecutive windows, which may produce fragmented and physiologically implausible sequences. To address this limitation, we propose a novel Temporal Structure Embedded Reinforcement Learning (TSE-RL) framework. This framework embeds prior temporal-structure knowledge of MI events into a stable RL formulation, reformulating window optimization as a constrained Markov Decision Process with ordinal relation enforcement of neural event. First, the evaluation and action selection upon each window has been changed from simply ``retain/discard'' to semantic alignment between window and neural events as \textit{Discard}/\textit{Onset}/\textit{sustain}/\textit{Offset}. Second, action masking is imposed by a finite-state machine (FSM) which enforces physiologically plausible decision orders with constraint of neural activity progress in term of  \textit{planning}/\textit{execution}/\textit{recovery}, enhancing both rationality, interpretability and classification performance. Third, a Transformer-based history encoder is designed to capture long-range inter-window dependencies in parallel, improving training stability and efficiency compared with existing recurrent architectures. Besides, TSE-RL adopts a plug-and-play design that can be integrated with diverse MI classifiers. Extensive experiments on multiple benchmark datasets demonstrate that TSE-RL consistently improves decoding performance in both cross-session and cross-subject settings. Visualization analyses further suggest that the learned policy of window optimization aligns with neurophysiological principles.
\end{abstract}


\section{Introduction}
\label{sec:introduction}
\IEEEPARstart{B}{rain} computer interface (BCI) technology establishes a direct communication and control pathway between the brain and external devices by decoding neural activity~\cite{clerc2013brain}. It has broad potential in diverse fields such as neurorehabilitation, entertainment~\cite{he2023diversity}, and human-machine interaction~\cite{liu2025novel}. Among non-invasive acquisition modalities, electroencephalography (EEG) is widely adopted due to its high temporal resolution and relatively low cost~\cite{vaid2015eeg}. In motor imagery (MI) BCIs, users voluntarily imagine limb movements to elicit event-related desynchronization/synchronization (ERD/ERS) patterns~\cite{ma2025decoding,wang2024eeg}, thereby enabling device control without external stimuli.

Recent years have witnessed significant advances in MI-EEG signal decoding techniques, most of which analyze a predefined post-cue segment (typically 0--4~s) for feature extraction and classification~\cite{altaheri2023deep}. Representative approaches range from Common Spatial Patterns (CSP)~\cite{lotte2010regularizing}, which learns spatial filters by maximizing inter-class variance, to diverse deep learning methods. Convolutional Neural Networks (CNNs) like EEGNet~\cite{lawhern2018eegnet} and M-ShallowConvNet~\cite{kim2022rethinking} efficiently capture spatio-temporal patterns with depthwise separable convolutions or shallow designs. FBCNet~\cite{mane2021fbcnet} performs frequency-band fusion via multi-view convolutions, while ATCNet~\cite{altaheri2022physics} integrates attention mechanisms with temporal convolutional networks. More recently, Transformer-based models have been explored in EEGConformer~\cite{song2022eeg} and TSLANet~\cite{eldele2024tslanet}. And CSPNet~\cite{jiang2024csp} embeds CSP principles within a neural network framework. Riemannian geometry-based methods such as manifold attention network (mAtt)~\cite{pan2022matt} characterize EEG patterns on the Riemannian manifold of symmetric positive-definite (SPD) matrices.

Although these methods perform well in offline evaluations, their performance in online scenarios are constrained by using a single, shared temporal segment to represent MI activity across trials.
For instance, online systems often relies on a sliding window of fixed length (around 1 s) moving along the continuous EEG signal. Some windows may cover only a fraction of an MI event, some may contain excessive pre-event baseline activity, and some may be contaminated by burst artifacts, the majority of which are misaligned with actual neural events. Meanwhile, fixed-length windows restrict the timeliness of the system response.
Online decoding thus can benefit from time window optimization, where windows are adaptively accumulated or discarded to form a segment that better matches informative neural patterns, providing better balance between timeliness and accuracy.

Researchers have proposed various time window optimization strategies.
Correlation-based Time Window Selection (CTWS)~\cite{feng2018towards} adjusts the
onset of window by correlating sliding windows with reference signals. Approaches
based on Ant Colony Optimization (ACO)~\cite{agustin2025optimized} and Sparrow
Search Algorithm (SSA)~\cite{huang2024adaptive} employ meta-learning and heuristic searches to identify optimal configurations in time-frequency parameter space. However, these
techniques typically evaluate and select candidate windows independently, treating
each as an isolated ``good/bad'' decision \rev{based on the quality of the window
, with configurations searched offline and applied uniformly across
trials}~\cite{wu2024motor, zhuo2024time}. This limitation can result in fragmented
or physiologically implausible window sequences that fail to capture the coherent
evolution of MI events. \rev{In contrast, our method optimizes the time window
sequence, conditioning each decision on cross-window history and on
the constraint of MI phases.} MI events exhibit contralateral
desynchronization of the $\mu$ (8--13 Hz) and $\beta$ (13--30 Hz) rhythms during MI
execution, and a subsequent rebound of synchronization of the $\beta$-rhythm after
the event ends~\cite{pfurtscheller1999event}. Such an evolution suggests distinct
MI phases from onset to sustained imagery and finally to offset.

Reinforcement Learning (RL) provides an ideal framework to overcome this limitation,
as it inherently models sequential decision-making processes with uncertainty. By
formulating the problem of time window optimization in MI-EEG as a Markov Decision
Process (MDP), an agent makes adaptive decisions at each time step based on the
current state and historical context (e.g., whether to retain a window and what
temporal role it plays in the MI event) to maximize cumulative rewards through
interactions with the environment. It also avoids brute-force search over window
combinations \rev{through exploration--exploitation}. In addition, Safe Reinforcement
Learning~\cite{gu2024review} also inspires us to incorporate neural event ordinal
information, where temporal-structure patterns must be embedded to constrain the
action space. Notably, preliminary studies have demonstrated the potential of RL in
improving BCI adaptivity. \rev{Aung et al.~\cite{aung2024eeg_rl} use RL to decide
whether to classify or skip the current time point, and Zhang et
al.~\cite{zhang2024attention} use RL to adaptively focus on features at varying scales. In
both, RL improves the adaptivity of the BCI decoding: the action
optimizes one component inside the decoding pipeline, while the analyzed window
remains fixed and predefined. Our work instead optimizes the window itself, taking
each incoming window as the decision unit, assigning it an explicit semantic role,
and constraining the action space by the structure of MI events.}

Therefore, we propose a Temporal Structure Embedded Reinforcement Learning (TSE-RL) framework for adaptive time window optimization in MI-EEG decoding. The core innovation of this framework is to embed prior temporal-structure knowledge of MI events into a stable RL framework, transforming window optimization from isolated ``good/bad'' judgments to semantic alignment of complete neural events. Specifically, TSE-RL incorporates three mechanisms to make this idea practical and reliable.
The first mechanism transforms the action selection upon each time window from simply ``retain/discard'' to semantic alignment between window and neural events as \textit{Discard}/\textit{Onset}/\textit{sustain}/\textit{Offset}. This semantic alignment mechanism imposes physiological patterns on window optimization of EEG signal.
The second mechanism adopts Finite State Machine (FSM) for constrained action masking which enables safe reinforcement learning in restricted action space. Without proper constraints, the RL agent may generate physiologically implausible window sequences that violate the known temporal structure of MI events (e.g., selecting an \textit{Offset} before an \textit{Onset}). The third mechanism concerns Transformer-based history encoding. The decisions of window optimization require integrating historical context, while traditional RL methods often employing Recurrent Neural Networks (RNNs) to capture temporal dependencies. When processing long sequences, RNNs are prone to gradient vanishing or explosion, resulting in training instability~\cite{osei2025deep}. The main contributions of this paper include:

\begin{itemize}
    \item We propose a temporal window optimization method that embeds MI event temporal structure knowledge into a reinforcement learning framework. The action space based on EEG time window evaluation is semantically aligned with neural event for interpretability and performance improvement. FSM-based action masking is designed to ensure physiological plausible decision sequence with constraint of neural activity progress in term of \textit{planning}/\textit{execution}/\textit{recovery}.
    \item We design a Transformer-based history encoder to replace training-unstable recurrent networks. This encoder takes historical window sequences as input and leverages self-attention mechanisms to learn temporal dependencies across windows in parallel, improving both training stability and computational efficiency.
    \item TSE-RL is plug-and-play and can be integrated with various mainstream MI classifiers. Simulated online experiments on multiple public datasets demonstrate that it significantly improves decoding performance in both cross-session and cross-subject settings.
\end{itemize}

\section{Methods}
\label{sec:method}

\begin{figure*}[!t]
    \centering
    \includegraphics[width=\textwidth]{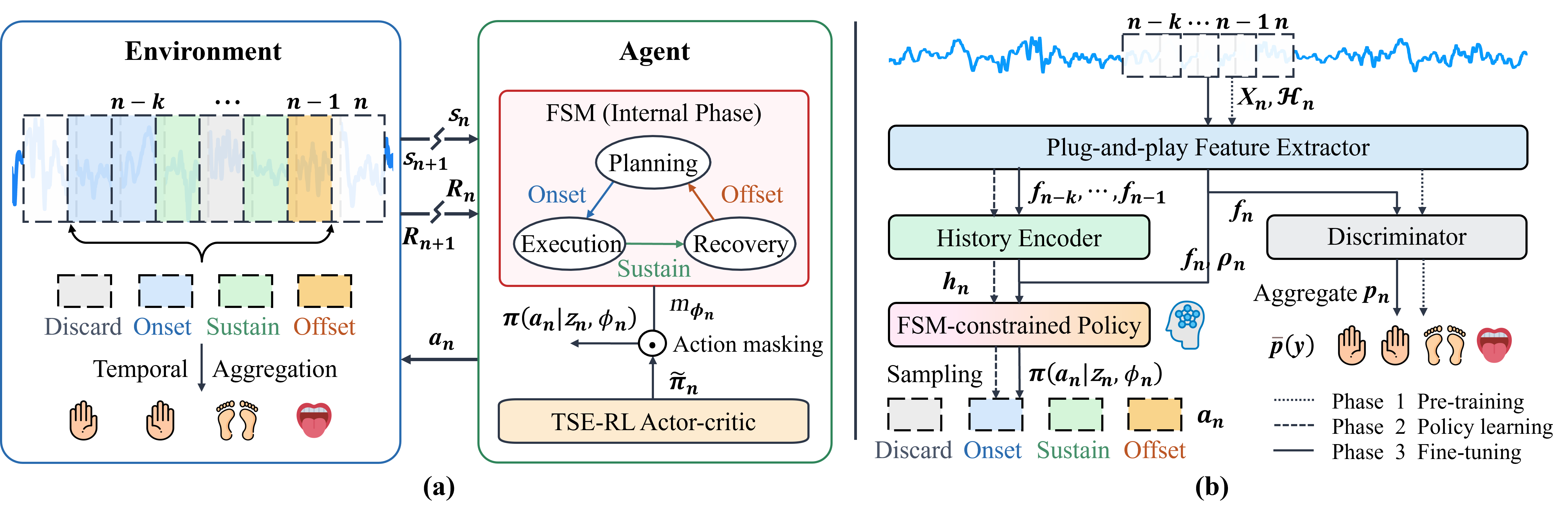}
    \caption{Overview of the proposed TSE-RL framework. (a) MDP formulation with an internal FSM for action masking. At step $n$, the agent receives the state $s_n$ from the environment. The FSM determines the valid action subset $\mathcal{A}_{\phi_n}$ for the current phase $\phi_n$ and constrains the masked policy $\tilde{\pi}_n$. The selected action $a_n$ drives the environment transition and the reward. (b) Network architecture of TSE-RL. The feature extractor encodes each window into a representation, and the discriminator outputs class probabilities for reward computation. A Transformer-based history encoder models temporal dependencies among historical features. The FSM-constrained policy network integrates current features, historical context, and trial progress to produce actions under phase-dependent masking.}

    \label{fig:overall}
\end{figure*}

We formulate time-window optimization for online MI-EEG decoding as an constrained sequential decision-making problem. In contrast to fixed-length window decoding and traditional window optimization, TSE-RL learns to assign each candidate window a semantic role (\textit{Discard}, \textit{Onset}, \textit{Sustain}, or \textit{Offset}).
As illustrated in Fig.~\ref{fig:overall}(a), TSE-RL encodes the MI event's temporal structure as a Finite State Machine (FSM) that governs the valid action space of an RL agent via action masking. This mechanism ensures the learned policy adheres to physiological plausibility, elevating window optimization from isolated ``good/bad'' judgments to semantic alignment of complete neural events. Moreover, TSE-RL employs a Transformer-based history encoder to stably integrate historical information in parallel, preventing the training instability and sequential overhead commonly observed in recurrent RL. This section presents the problem formulation, the temporal-structure embedding mechanism, the overall framework, and the learning procedure.

\subsection{Problem Formulation}
\label{subsec:problem_formulation}

Given a labeled EEG dataset $\mathcal{D}=\{(X,y)\}$, where $y\in\mathcal{L}$ denotes the class label and $X=[x_1,\cdots,x_T]\in\mathbb{R}^{C\times T}$ represents a trial of length $T\in\mathbb{N}^+$ acquired from $C\in\mathbb{N}^+$ electrodes, with $x_t\in\mathbb{R}^C$ denoting the signal sample recorded at time step $t\in[1,T]$. Online decoding methods employ sliding windows for continuous trials, with window length $L$ and stride $\Delta$. The $n$-th window is denoted as $X_n=X_{n\Delta:(n\Delta+L)}\in\mathbb{R}^{C\times L}$, where $n=0, 1, \cdots, \lfloor\frac{T-L}{\Delta}\rfloor$. We want to learn an optimal policy $\pi^\ast$ that can be utilised during the processing of the window sequence $\{X_n\}$. At each time step, this policy dynamically decides whether the current window should be discarded or retained. And if retained, the policy must decide whether the window should be marked as the \emph{Onset}, \emph{Sustain}, or \emph{Offset} phase of the MI event. The objective is to retain and mark an sequence of most informative windows such that the final classification accuracy based on this sequence is maximized. We formulate this problem as a Markov Decision Process (MDP) defined by the tuple $(\mathcal{S},\mathcal{A},\mathcal{T},R,\gamma)$.

\textbf{State set $\mathcal{S}$:} At each time step $n$, the environment state is defined as $s_n=(X_n, \mathcal{H}_n, \rho_n)$, where $X_n$ is the current window, $\mathcal{H}_n=\{X_{n-k},\dots,X_{n-1}\}$ denotes fixed-length historical windows (with $k$ being the history length), and $\rho_n=\frac{n\Delta}{T-L}\in[0,1]$ is a normalized indicator of trial progress. In the implementation, $X_n$ and $\mathcal{H}_n$ are further encoded into feature vectors by neural networks, but conceptually the MDP state is characterized by $(X_n,\mathcal{H}_n,\rho_n)$ throughout this paper.

\textbf{Action set $\mathcal{A}$:} The agent executes one of the four discrete actions $a_n \in \{0, 1, 2, 3\}$. Specifically, action $a_n=0$ indicates that the current window is considered to be highly noisy and is discarded. Action $a_n=1$ denotes that the current window is retained and marked as the \emph{Onset} phase of the MI event. Action $a_n=2$ denotes the current window is retained and marked as the \emph{Sustain} phase. Action $a_n=3$ denotes the current window is retained and marked as the \emph{Offset} phase. When $a_n=3$ is taken, or when the sliding window reaches the end of the trial, the agent terminates the current trial and triggers the final classification based on the sequence of the retained windows.

\textbf{State transition function $\mathcal{T}$:} After the agent selects $a_n$, the environment either advances the window to $X_{n+1}$ and updates $(\mathcal{H}_{n+1},\rho_{n+1})$, or terminates the trial. To incorporate the prior knowledge that an MI event follows an \emph{Onset}$\rightarrow$\emph{Sustain}$\rightarrow$\emph{Offset} temporal structure, we introduce a discrete phase indicator $\phi_n \in \{\textit{Planning},\textit{Execution},\textit{Recovery}\}$. Importantly, $\phi_n$ is an internal phase of sensorimotor cortex rather than an additional MDP state. It is only used to constrain the legal actions given the current phase via a finite-state machine (FSM).

\begin{itemize}
    \item \textit{Planning} phase. At the beginning of each trial, the agent is in the \textit{Planning} phase. It has been demonstrated that subtle rhythmic changes often precede the actual onset of MI performing, representing a preparation or planning phase ~\cite{pfurtscheller1999event}.
    In this phase, the agent searches for the onset of the event, and the legal actions are
    $
        \mathcal{A}_{\textit{Planning}} = \{0\ (\text{discard}),\,1\ (\text{Onset})\}.
    $
    Choosing $a_n=1$ marks the current window as Onset and switches the phase to \textit{Execution}.

    \item \textit{Execution} phase. After the first Onset has been detected, the agent enters the \textit{Execution} phase, which covers the duration of the active MI. During this phase, the sensorimotor cortex typically exhibits a significant power reduction in the $\mu$ (8--13 Hz) and $\beta$ (13--30 Hz) rhythms, a phenomenon known as ERD~\cite{pfurtscheller1999event}. This reduction persists as long as the subject actively performs the imagery task.
    Accordingly, the legal actions are
    $
        \mathcal{A}_{\textit{Execution}} = \{0,\,1,\,2\ (\text{Sustain})\}.
    $
    The agent may refine the onset boundary (action 1) or, upon confirming stable ERD patterns, select $a_n=2$ to capture the sustained activity. Selecting $a_n=2$ triggers the transition to the \textit{Recovery} phase.

    \item \textit{Recovery} phase. After a sustained segment is established, the agent enters the \textit{Recovery} phase. This corresponds to the post-MI phase where sensorimotor rhythms return to baseline. Specifically, a distinct power rebound in the $\beta$ rhythms often occurs after MI performing, termed ERS~\cite{pfurtscheller1999event}.
    To accommodate this, the legal actions are
    $
        \mathcal{A}_{\textit{Recovery}} = \{0,\,2,\,3\ (\text{Offset})\}.
    $
    The agent can continue to extend the sustained segment (action 2) to cover late ERD components or, upon detecting the ERS signature, select $a_n=3$ to mark the event termination. This action concludes the trial and resets the phase to \textit{Planning} for the subsequent trial.
\end{itemize}

This FSM therefore implements the following intuitive rules: the agent must first identify the preparatory \emph{Onset} during the \textit{Planning} phase, then capture the sustained ERD during the \textit{Execution} phase, and finally determine the \emph{Offset} based on ERS patterns during the \textit{Recovery} phase.

\textbf{Reward function $R$:} The reward function is defined as $R = r_{\text{disc}}(a_n) + r_{\text{cont}}(a_{n-1}, a_n) + r_{\text{term}}(a_n)$, where the discrimination reward $r_{\text{disc}}$ provides immediate feedback on the quality of the current window, the temporal continuity reward $r_{\text{cont}}$ encourages smooth window sequences, and the termination reward $r_{\text{term}}$ reflects the final classification outcome of the retained windows.

The discrimination reward $r_{\text{disc}}$ provides immediate feedback based on the classifier's confidence $p_n(y_{\text{true}})$ for the current window, encouraging retention of informative windows while discouraging ambiguous ones.
\begin{equation}
\label{eq:r_disc}
r_{\text{disc}}(a_n) =
\begin{cases}
2 \cdot (p_n(y_{\text{true}}) - 0.5), & \text{if } a_n \in \{1,2,3\}, \\
0.2 \cdot (0.5 - p_n(y_{\text{true}})), & \text{if } a_n = 0.
\end{cases}
\end{equation}

The continuity reward $r_{\text{cont}}$ promotes consistent decisions (e.g., Sustain followed by Sustain), and penalizes unnecessary discards once the agent has started to track an MI segment.

\begin{equation}
\label{eq:r_cont}
\begin{aligned}
r_{\text{cont}}(a_{n-1}, a_n)
&= 0.1\,\mathbb{I}_{\!\left\{a_n>0 \,\land\, a_n=a_{n-1}\right\}} \\
&\quad - 0.1\,\mathbb{I}_{\!\left\{\phi_n \neq \textit{Planning} \,\land\, a_n=0\right\}},
\end{aligned}
\end{equation}
where $\phi_n$ is the current FSM phase indicator and $\mathbb{I}_{\{\cdot\}}$ is the indicator function.

The termination reward $r_{\text{term}}$ is issued when the trial ends, either
via an Offset action ($a_n=3$) or when the sliding window exceeds the trial
length. Let $\mathcal{I}=\{i \mid a_i \neq 0\}$ index the retained windows. The
retained windows are aggregated into a trial-level class distribution $\bar{p}$
by a weighted average of window-wise class probabilities
\begin{equation}
\label{eq:weighted_avg}
\bar{p}(y) = \frac{\sum_{i\in\mathcal{I}} w(a_i)\, p_i(y)}{\sum_{i\in\mathcal{I}} w(a_i)},
\end{equation}
where $w(1)=1.0$, $w(2)=1.5$, and $w(3)=1.0$. \rev{Sustain windows cover the stage of the event where the ERD is fully
developed, and are weighted highest. Onset and Offset windows lie at its
boundaries, where the ERD is developing or returning to
baseline, and are weighted lower.} If $\mathcal{I}=\varnothing$, we set
$\bar{p}(y)=1/|\mathcal{L}|$ for all $y\in\mathcal{L}$. The final prediction is
$\hat{y}=\arg\max_{y} \bar{p}(y)$, and the termination reward is
\begin{equation}
\label{eq:r_term}
r_{\text{term}}(a_n) =
\begin{cases}
5 \cdot \alpha(N), & \text{if } \hat{y} = y_{\text{true}}, \\
-5 \cdot \alpha(N), & \text{otherwise},
\end{cases}
\end{equation}
where $N = |\mathcal{I}|$ is the number of retained windows and
$\alpha(N)=1 + 0.1\cdot\max(0,N-3)$ is a length-dependent multiplier.
\rev{Both $r_{\text{disc}}$ and $r_{\text{term}}$ are evaluated from the
probabilities produced by $D$, as public MI datasets provide trial-level labels
only. The temporal structure of MI events is thus imposed not by the reward
alone but jointly with the FSM-constrained action mask of Section~II-B.}

\textbf{Discount factor $\gamma$:} $\gamma=0.99$ is used so that the agent
optimizes cumulative long-term returns.

\subsection{Temporal structure embedding via FSM-constrained action masking and Transformer-based history encoding}

As mentioned in Introduction, applying RL to MI-EEG time window optimization faces two significant challenges. First, the agent must respect safety constraints so that selected window sequences remain physiologically meaningful. Second, the agent has to learn from long sequences of noisy EEG observations in a stable and data-efficient way. The proposed TSE-RL framework addresses these issues through a temporal structure embedding mechanism that combines an FSM-constrained action mask with a Transformer-based history encoder. This design enforces the decision of window optimization alignment with MI neural patterns, while providing a stable architecture for learning.

MI events exhibit contralateral desynchronization of the $\mu$ (8–13 Hz) and $\beta$ (13–30 Hz) rhythms during MI execution, and exhibit a rebound of synchronization of the $\beta$-rhythm after the event ends~\cite{pfurtscheller1999event}. Conventional isolated window optimization methods ignore this temporal pattern, leading to discontinuous or semantically incomplete window sequences. This consequently reduces classification accuracy and diminishes interpretability. Reward shaping can, in principle, discourage such sequences by penalizing non-contiguous or incomplete window selections. However, this soft constraint can only indirectly influence the policy through policy gradients. During the early stages of training, the policy that is not aligned with physiological patterns can still be explored. Even during inference, a converged policy employing stochastic sampling cannot entirely prevent the occurrence of invalid action sequences.

To ensure that the selected time window consistently constitutes a valid MI event, we embed the temporal structure of MI events into the decision process using FSM-constrained action masking. We employ FSM to model the legal phase transitions of events, restricting valid actions $\mathcal{A}_{\phi_n}$ based on the current phase $\phi_n$. The probability of all invalid actions is set to zero before the action sampling.
This hard constraint reduces the policy search space, forcing the agent to follow physiologically plausible trajectories. Huang et al.~\cite{huang2022invalid} showed that invalid-action masking yields more effective gradient updates, with benefits increasing as the number of invalid actions grows. Masking scales well when invalid actions are numerous, while penalty-based methods struggle to explore the task. Furthermore, Stolz et al.~\cite{stolz2024excluding} found that focusing exploration on state-specific valid actions significantly improves learning efficiency and the predictability of the agent in safety‑critical tasks. Therefore, in EEG decoding, we propose FSM-constrained action masking combined with the reward function described in Section~\ref{subsec:problem_formulation} to guide the agent to generate continuous and complete MI windows.

The second challenge is to integrate information over the histories of extended observation when the agent determines actions. Traditional RL methods employ RNNs, but they suffer from instability, particularly in MI-EEG tasks with trials spanning dozens of windows over several seconds~\cite{osei2025deep}.
Each additional window in the sequence elongates the gradient path, causing gradient to decay or explode exponentially. Beyond instability, RNNs require careful management of variable-length sequences and hidden states during training, complicating the implementation of batching strategies. These challenges can hinder the deployment of recurrent RL approaches in MI-EEG scenarios.

We thereby adopts a non-recurrent Transformer-based history encoder. Instead of propagating hidden states, it processes the recent history $\mathcal{H}_n$ (as feature vectors $\{f_{n-k}, \dots, f_{n-1}\}$) in parallel using self-attention mechanisms.
This generates a fixed-dimensional context vector $h_n$ that captures temporal dependencies across history observations. When concatenated with the current window feature $f_n$ and trial progress $\rho_n$, this forms the full input for the policy network. By eliminating recursive propagation of hidden state, the design circumvents gradient instability while leveraging parallel computation for efficiency. The Transformer's multi-head attention effectively captures long-range temporal dependencies. This architecture has demonstrated superior performance in partial-observation RL tasks~\cite{wang2025datd3}.

\subsection{Overall TSE-RL framework}
\label{subsec:history_encoder}

The TSE-RL framework is established to solve the MI-EEG time window optimization MDP, where the agent learns an adaptive policy to decide whether to discard, start, sustain or terminate a MI segment at each time step. As illustrated in Fig.~\ref{fig:overall}(b), a feature extractor $F$ and a discriminator $D$ is used for window-wise assessment which is decomposed from a plug-and-play EEG backbone; a Transformer-based history encoder $G$ summarizes historical features; and an FSM-constrained policy network $\Pi$ outputs actions under the phase-dependent action mask.

\textbf{Plug-and-play feature extractor and discriminator.}
Let $F:\mathbb{R}^{C\times L}\rightarrow\mathbb{R}^d$ denote the feature extractor and $D:\mathbb{R}^d\rightarrow\mathbb{R}^{|\mathcal{L}|}$ denote the discriminator. Given an established EEG classifier $B$ that maps an input window to class probabilities, we can write it as
\begin{equation}
    B(X_n) = D\bigl(F(X_n)\bigr),
\end{equation}
where $F$ consists of the backbone's feature extraction layers and $D$ consists of the remaining layers plus the final classification head. It is important to note that this decomposition is applicable to a wide range of EEG backbones, including, but not limited to, convolutional, recurrent, and Transformer-based EEG networks.

For each incoming window $X_n$, the extractor produces a feature vector
\begin{equation}
    f_n = F(X_n) \in \mathbb{R}^d,
\end{equation}
which is fed into the discriminator to obtain the probability distribution of class per window, i.e.,
\begin{equation}
    p_n = D(f_n) \in \mathbb{R}^{|\mathcal{L}|}.
\end{equation}

\rev{For a given dataset, $L$ and $\Delta$ are fixed. Actions determine only whether a window is retained and which semantic role it takes.
Thus, the extractor therefore always operates on windows of the same size.
Fixed-dimensional class probabilities are aggregated by
\eqref{eq:weighted_avg}, so that accumulated EEG segments of different lengths can be classified without adjusting the input size.}

\begin{table*}[t]
\centering
\small

\setlength{\tabcolsep}{8pt}
\caption{Summary of the \rev{four} MI-EEG datasets}
\label{tab:eeg_datasets}
\begin{tabular}{@{}lccccccc@{}}
\toprule
Dataset &
\makecell[c]{Number of\\subjects} &
\makecell[c]{Number of\\channels} &
\makecell[c]{Sampling\\rate (Hz)} &
\makecell[c]{Trial\\length (s)} &
\makecell[c]{Number of\\sessions} &
\makecell[c]{Trials in\\1st session} &
\makecell[c]{Types of\\imaginations} \\
\midrule
BNCI2014001 & 9  & 22 & 250 & 4 & 2 & 288 & r/l hand, feet, tongue \\
BNCI2014002 & 14 & 15 & 512 & 5 & 1 & 160 & right hand, both feet \\
BNCI2015001 & 12 & 13 & 512 & 5 & 3 & 400 & right hand, both feet \\
\rev{OpenBMI} & \rev{54} & \rev{62} & \rev{1000} & \rev{4} & \rev{2} & \rev{200} & \rev{right/left hand } \\
\bottomrule
\end{tabular}
\end{table*}

\textbf{History encoder.}
To capture temporal dependencies, the history encoder $G$ operates on the features of the recent window buffer $\mathcal{H}_n=\{X_{n-k},\dots,X_{n-1}\}$. The application of extractor $F$ to each historical window yields a feature sequence
\begin{equation}
    (f_{n-k},\dots,f_{n-1}) = \bigl(F(X_{n-k}),\dots,F(X_{n-1})\bigr).
\end{equation}
These features are then fed into a standard Transformer encoder $G$. The encoder computes self-attention across the sequence, aggregating contextual information. This is followed by mean pooling over the temporal dimension, which is then succeeded by a linear projection to a fixed-dimensional context vector
\begin{equation}
    h_n = G(f_{n-k},\dots,f_{n-1}) \in \mathbb{R}^{d_h}.
\end{equation}
Here $d_h$ denotes the context dimension. This design yields a stable, fixed-dimensional representation of variable-length histories without relying on recurrent computations.

\textbf{FSM-constrained policy network.}
At each time step $n$, we construct the policy input vector by concatenating the current feature, the historical context, and the trial progress indicator, i.e. $z_n = [f_n;\,h_n;\,\rho_n] \in \mathbb{R}^{d+d_h+1}$.
The FSM-constrained policy network $\Pi$ implements an actor--critic architecture on top of $z_n$. A shared multilayer perceptron first transforms $z_n$ into a latent representation, which then branches into an \emph{actor} head producing action logits $\ell_n \in \mathbb{R}^{|\mathcal{A}|}$ and a \emph{critic} head producing a scalar value estimate $V_n\in\mathbb{R}$ for the current state.
The critic is trained jointly with the actor using Proximal Policy Optimization, but for notational simplicity we focus on the actor branch in the following computation graph.

Let $\tilde{\pi}_n = \mathrm{softmax}(\ell_n)$ denote the unconstrained action probability vector output by the actor head. The current FSM phase $\phi_n$ (defined in Section~\ref{subsec:problem_formulation}) determines a binary action mask $m_{\phi_n} \in \{0,1\}^{|\mathcal{A}|}$, which encodes the legal actions in that phase. The final action distribution used for sampling is obtained by masking and renormalization:
\begin{equation}
\label{eq:5}
\begin{aligned}
\tilde{\pi}_n &= \mathrm{softmax}\bigl(\Pi(z_n)\bigr), \\
\pi(a_n \mid z_n,\phi_n) &\propto \tilde{\pi}_n \odot m_{\phi_n}, \\
a_n &\sim \pi(\cdot \mid z_n,\phi_n),
\end{aligned}
\end{equation}
where $\odot$ denotes element-wise multiplication and the $\propto$ indicates that the masked probabilities are renormalized to sum to one. In this way, the policy is hard constrained by the FSM. Actions that violate the supported Onset--Sustain--Offset pattern are assigned zero probability and never sampled during either training or inference.

Overall, the feature extractor $F$ provides a consistent representational feature, the discriminator $D$ provides discrimination and terminal rewards, the history encoder $G$ captures temporal context, and the FSM-constrained policy network $\Pi$ integrates these information to produce window-level decisions that respect the MI event structure encoded by the FSM.

\subsection{Learning framework}
\label{subsec:learning_framework}

To address the low sample efficiency and training instability of end-to-end RL from random initialization, the TSE-RL framework is trained in three stages that progressively shift from supervised learning to policy optimization.

\textbf{Stage 1: Supervised pretraining.}
In the first stage, we train the feature extractor $F$ and discriminator $D$ as a standard classifier on full time windows. From each training trial, we extract time windows $X_n$ using the window length $L$ and stride $\Delta$, and assign each window the corresponding trial label $y_{\text{true}}$. The backbone then produces $p_n = D\bigl(F(X_n)\bigr)$ for each window,
and we minimize the cross-entropy loss
\begin{equation}
    \mathcal{L}_{\text{CE}} = - \sum_{n} \log p_n(y_{\text{true}}),
\end{equation}
temporally aggregated over all windows in a mini-batch. Trial-level validation accuracy can be computed by averaging window-level probabilities. At the end of this stage, $F$ has learned discriminative spatio-temporal representations of EEG windows, and $D$ has learned the class distribution whose probabilities $p_n$ will be reused in the reward design.

\textbf{Stage 2: Policy learning.}
After the convergence of supervised pretraining, the parameters of $F$ and $D$ are frozen. Then we train the history encoder $G$ and the FSM-constrained policy network $\Pi$ by reinforcement learning. We adopt Proximal Policy Optimization (PPO)~\cite{schulman2017proximal} algorithm. In each trial, the agent interacts with the environment as described in Section~\ref{subsec:problem_formulation}: the frozen extractor $F$ converts the current and historical windows into features, the frozen discriminator $D$ provides window-level probabilities $p_n$ for computing $r_{\text{disc}}$ and $r_{\text{term}}$, the history encoder $G$ produces the context vector $h_n$, and the policy network $\Pi$ maps $z_n = [f_n;h_n;\rho_n]$ to an action distribution that is masked by the current phase $\phi_n$.

During this stage, PPO gradients update only the parameters of $G$ and $\Pi$. Consequently, the mapping from raw windows to features and probabilities does not change while the agent is learning decision policy. This separation prevents representation drift from interfering with policy optimization, making training more stable and data-efficient. The FSM-based action mask is active throughout, restricting the exploration to physiologically plausible action.

\textbf{Stage 3: Joint fine-tuning.}
Once the policy has reached stable performance, we perform a final joint fine-tuning. We unfreeze the last few layers of the feature extractor $F$ while keeping the discriminator $D$ fixed, and continue PPO training with a smaller learning rate. Policy gradients are now allowed to slightly adjust the high-level feature representations, which allows mild adaptation of the feature representation to the learned policy while preserving the stability gained in Stage~2.

\rev{The pseudocode of the complete pipeline is given as Algorithm~1 in the
Supplementary Material. The source
code and the Supplementary Material are available at \url{https://github.com/LunaLab-XJTU/TSE-RL}.}

\begin{table*}[!t]
\centering
\small

\setlength{\tabcolsep}{6pt}
\caption{Within-subject classification accuracy (\%) in simulated online experiment on BNCI2014001, with an asterisk (*) denoting the significance level (*: p<0.05).}
\label{tab:BNCI2014001_within}
\begin{tabular}{@{}L{1cm} l | ccccccccc | c@{}}
\toprule
\textbf{Setting} & \textbf{Method} & \textbf{S0} & \textbf{S1} & \textbf{S2} & \textbf{S3} & \textbf{S4} & \textbf{S5} & \textbf{S6} & \textbf{S7} & \textbf{S8} & \textbf{Avg.} \\
\midrule
\multirow{8}{1.1cm}{\raggedright Baseline-full}
 & EEGNet        & 84.97 & 53.51 & 86.15 & 61.92 & 54.14 & 52.54 & 76.85 & 80.11 & 81.16 & $70.15_{\pm 2.04}$ \\
 & TSLANet       & 88.38 & 51.22 & 84.56 & 47.40 & 37.68 & 58.86 & 77.61 & 84.56 & 82.12 & $\mathbf{68.04}_{\pm 2.43}$ \\
 & EEGConformer  & 84.51 & 49.83 & 89.19 & 63.41 & 46.57 & 55.86 & 86.50 & 79.30 & 81.73 & $70.77_{\pm 0.92}$ \\
 & FBCNet        & 82.33 & 58.38 & 81.64 & 68.79 & 44.14 & 51.78 & 84.07 & 83.38 & 84.07 & $70.95_{\pm 1.58}$ \\
 & ATCNet        & 87.26 & 54.56 & 89.21 & 65.95 & 57.26 & 59.07 & 82.96 & 83.72 & 83.65 & $73.74_{\pm 1.33}$ \\
 & M-ShallowConvNet  & 86.50 & 54.84 & 91.50 & 70.94 & 57.47 & 58.86 & 89.07 & 83.51 & 83.65 & $75.15_{\pm 1.21}$ \\
 & mAtt          & 87.15 & 58.68 & 83.68 & 64.93 & 51.74 & 47.57 & 79.86 & 83.68 & 85.07 & $71.37_{\pm 1.66}$ \\
 & CSPNet        & 88.24 & 57.68 & 88.93 & 70.53 & 62.19 & 60.81 & 87.89 & 82.68 & 84.07 & $75.89_{\pm 1.34}$ \\
\midrule
\multirow{8}{1.4cm}{\raggedright TSE-RL}
 & EEGNet        & 86.59 & 54.91 & 88.82 & 63.68 & 54.93 & 56.39 & 79.38 & 82.78 & 83.75 & $\mathbf{72.34^{*}_{\pm 1.91}}$ \\
 & TSLANet       & 88.72 & 48.44 & 76.56 & 46.71 & 37.68 & 56.78 & 74.49 & 86.99 & 83.51 & $66.65_{\pm 2.09}$ \\
 & EEGConformer  & 86.54 & 47.66 & 91.23 & 64.24 & 47.48 & 55.82 & 87.59 & 80.38 & 82.81 & $\mathbf{71.52}_{\pm 1.85}$ \\
 & FBCNet        & 83.33 & 61.81 & 85.07 & 70.14 & 47.57 & 53.13 & 86.11 & 86.11 & 87.85 & $\mathbf{73.46^{*}_{\pm 1.89}}$ \\
 & ATCNet        & 88.47 & 56.32 & 90.97 & 66.74 & 59.51 & 61.32 & 85.07 & 85.76 & 85.00 & $\mathbf{75.46^{*}_{\pm 1.29}}$ \\
 & M-ShallowConvNet  & 88.06 & 56.94 & 92.78 & 72.08 & 60.42 & 62.92 & 91.46 & 86.04 & 84.93 & $\mathbf{77.29^{*}_{\pm 1.31}}$ \\
 & mAtt          & 87.50 & 57.99 & 84.72 & 66.67 & 53.47 & 49.31 & 79.86 & 83.68 & 85.42 & $\mathbf{72.07}_{\pm 1.56}$ \\
 & CSPNet        & 89.58 & 61.46 & 90.63 & 73.26 & 66.32 & 63.19 & 89.24 & 86.11 & 86.46 & $\mathbf{78.47^{*}_{\pm 1.25}}$ \\
\bottomrule
\end{tabular}
\end{table*}

\begin{table}[!t]
\centering
\footnotesize
\setlength{\tabcolsep}{3pt}
\renewcommand{\arraystretch}{1.0}
\caption{Average within-subject classification accuracy (\%) in simulated online experiments on BNCI2014002, BNCI2015001 \rev{and OpenBMI}. * indicates $p<0.05$.}
\label{tab:within_summary_4002_5001}
\begin{tabular}{@{}p{1cm} l ccc@{}}
\toprule
\textbf{Setting} & \textbf{Method} & \textbf{BNCI2014002} & \textbf{BNCI2015001} & \rev{\textbf{OpenBMI}} \\
\midrule
\multirow{8}{1.1cm}{\raggedright Baseline-full}
 & EEGNet            & $85.59_{\pm 1.91}$ & $82.42_{\pm 1.96}$ & \rev{$71.97_{\pm 1.30}$} \\
 & TSLANet           & $83.33_{\pm 2.80}$ & $85.79_{\pm 2.64}$ & \rev{$73.67_{\pm 1.65}$} \\
 & EEGConformer      & $76.83_{\pm 1.54}$ & $85.29_{\pm 1.67}$ & \rev{$73.94_{\pm 1.31}$} \\
 & FBCNet            & $87.26_{\pm 1.56}$ & $83.91_{\pm 1.53}$ & \rev{$74.93_{\pm 1.09}$} \\
 & ATCNet            & $88.81_{\pm 1.25}$ & $87.96_{\pm 1.29}$ & \rev{$75.46_{\pm 1.35}$} \\
 & M-ShallowConvNet  & $88.73_{\pm 1.27}$ & $86.37_{\pm 1.14}$ & \rev{$74.67_{\pm 1.48}$} \\
 & mAtt              & $84.52_{\pm 1.74}$ & $82.87_{\pm 1.78}$ & \rev{$72.83_{\pm 1.05}$} \\
 & CSPNet            & $84.92_{\pm 1.37}$ & $85.37_{\pm 1.49}$ & \rev{$73.15_{\pm 1.25}$} \\
\midrule
\multirow{8}{1.4cm}{\raggedright TSE-RL}
 & EEGNet            & $\mathbf{87.62^{*}_{\pm 1.67}}$ & $\mathbf{84.88^{*}_{\pm 1.71}}$ & \rev{$\mathbf{74.70^{*}_{\pm 1.64}}$} \\
 & TSLANet           & $\mathbf{84.40_{\pm 2.68}}$     & $\mathbf{85.92_{\pm 2.35}}$     & \rev{$\mathbf{76.22^{*}_{\pm 1.72}}$} \\
 & EEGConformer      & $\mathbf{79.05^{*}_{\pm 1.97}}$ & $\mathbf{87.21^{*}_{\pm 1.90}}$ & \rev{$\mathbf{76.64^{*}_{\pm 1.74}}$} \\
 & FBCNet            & $\mathbf{87.98_{\pm 1.62}}$     & $\mathbf{85.29_{\pm 1.45}}$     & \rev{$\mathbf{76.79_{\pm 1.74}}$} \\
 & ATCNet            & $\mathbf{90.95^{*}_{\pm 1.54}}$ & $\mathbf{89.08_{\pm 1.58}}$     & \rev{$\mathbf{77.36_{\pm 1.83}}$} \\
 & M-ShallowConvNet  & $\mathbf{90.95^{*}_{\pm 1.36}}$ & $\mathbf{88.38^{*}_{\pm 1.10}}$ & \rev{$\mathbf{77.02^{*}_{\pm 1.79}}$} \\
 & mAtt              & $\mathbf{85.83_{\pm 1.66}}$     & $\mathbf{84.54_{\pm 1.71}}$     & \rev{$\mathbf{74.43_{\pm 1.88}}$} \\
 & CSPNet            & $\mathbf{86.91^{*}_{\pm 1.22}}$ & $\mathbf{87.48^{*}_{\pm 1.27}}$ & \rev{$\mathbf{75.50^{*}_{\pm 1.69}}$} \\
\bottomrule
\end{tabular}
\end{table}

\section{Experimental results and discussion}
\label{sec:results}

\subsection{Datasets and preprocessing}
\label{subsec:datasets}

TSE-RL is evaluated on \rev{four} public MI-EEG datasets from MOABB~\cite{jayaram2018moabb},
namely BNCI2014001, BNCI2014002, \rev{BNCI2015001 and OpenBMI~\cite{lee2019openbmi}}. Table~\ref{tab:eeg_datasets} summarizes their statistics. BNCI2014001 is derived from BCI Competition IV-2a~\cite{tangermann2012review} and is the only four-class dataset, while the other three are two-class. Two experimental protocols are considered. For within-subject cross-session experiments, the T session of BNCI2014001 is used for training, while the E session is used for testing. For BNCI2014002, the first five runs are used for training, with the remaining three runs used for testing. In the case of BNCI2015001, the first session is used for training, while the second
session is used for testing\rev{, and the same split is applied to OpenBMI}.
For cross-subject experiments, a leave-one-subject-out (LOSO) protocol is adopted in which, for each fold, a single subject serves as the test set and all remaining subjects are merged to form the training set.

All datasets share the same preprocessing steps. Raw EEG signals are band-pass filtered
(8--32~Hz), then multiplied by a factor of $10^6$ to scale to microvolts ($\mu$V). For each
recording block used in training or testing (session or run, depending on the dataset), all
trials are concatenated along the time axis\rev{, in their original recording order,} to form
a continuous EEG dataflow that simulates online decoding. Sliding windows with length
$L = 1.0$~s and stride $\Delta = 0.1$~s are applied to this EEG dataflow, producing a sequence
of time windows $\{X_n\}$ as defined in Section~\ref{subsec:problem_formulation}. At each time
step, the TSE-RL agent interacts with this unlabeled window sequence and decides whether to
discard the current window or to retain it as the Onset, Sustain, or Offset phase. \rev{The
true trial boundaries and labels are used only to compute rewards during training and to
evaluate accuracy during testing, and remain inaccessible to the agent. Decisions are  causal in that the agent
observes only the current window $X_n$ and the finite history $\mathcal{H}_n$ at step $n$,
while no future sample is available.
At test time the trials are decoded one by one with a batch size of one, and no
batch-normalization statistics are computed across trials, which rules out any leakage of
information between trials.}

\subsection{Implementation details}
\label{subsec:impl}

\textbf{Baselines.}
Eight representative MI-EEG classification backbones are used as supervised baselines and backbones for TSE-RL: EEGNet~\cite{lawhern2018eegnet}, M-ShallowConvNet~\cite{kim2022rethinking}, FBCNet~\cite{mane2021fbcnet}, ATCNet~\cite{altaheri2022physics}, EEGConformer~\cite{song2022eeg}, TSLANet~\cite{eldele2024tslanet}, CSPNet~\cite{jiang2024csp} and mAtt~\cite{pan2022matt}. Hyperparameters of these backbones follow the recommendations in the original papers or the official code repositories.
For each backbone $B$, the network is decomposed as $B(X_n)=D(F(X_n))$ following the principle in Section~\ref{subsec:history_encoder}.
Concretely, the split between $F$ and $D$ is defined as follows:
\begin{itemize}
  \item \textbf{EEGNet}~\cite{lawhern2018eegnet}: $F$ contains the initial temporal convolution and depthwise spatial convolution blocks (including batch normalization, ELU, and average pooling), which produce a compact spatio–temporal feature map. $D$ comprises the final separable convolution, global average pooling, dropout, and the dense classification layer that maps the pooled feature to class logits.

  \item \textbf{M-ShallowConvNet}~\cite{kim2022rethinking}: $F$ contains the temporal convolution and depthwise spatial convolution layers together with the first batch normalization. $D$ includes the squaring non-linearity, mean pooling, log transformation, dropout, and the final fully connected classifier.

  \item \textbf{FBCNet}~\cite{mane2021fbcnet}: $F$ is the FBCNet feature extractor that applies multi-band temporal convolutions and spatial filters for each filter bank branch and outputs concatenated feature maps across bands. $D$ is the FBCNet discriminator that performs covariance-based pooling over the spatially filtered signals and projects the pooled representation to class logits through fully connected layers.

  \item \textbf{ATCNet}~\cite{altaheri2022physics}: $F$ consists of the temporal, depthwise and spatial convolution layers, and the stacked attentional temporal convolution blocks that form the main encoder. $D$ contains the final global average pooling, dropout, and classification layers that operate on the encoder output.

  \item \textbf{EEGConformer}~\cite{song2022eeg}: $F$ is the temporal and depthwise convolution layers. $D$ includes the remaining convolution layers, Conformer encoder, and the output linear classifier.

  \item \textbf{TSLANet}~\cite{eldele2024tslanet}: $F$ comprises the temporal segmentation and patch embedding, and positional encoding. $D$ consists of the stacked temporal–spectral local attention blocks, final pooling over the sequence dimension and the dense classification head.

  \item \textbf{CSPNet}~\cite{jiang2024csp}: $F$ includes the filter-bank temporal convolutions and CSP-like spatial projection layers that produce multi-band spatial features. $D$ aggregates these features via pooling and passes them through the final fully connected classifier.

  \item \textbf{mAtt}~\cite{pan2022matt}: $F$ is the multi-head Riemannian feature extractor that computes, for each window, covariance matrices, projects them to a low-dimensional subspace, and applies the matrix logarithm to obtain a stack of normalized log-SPD matrices without flattening. $D$ vectorizes the per-head SPD features, applies an attention mechanism to weight and aggregate the heads, and feeds the aggregated representation into an MLP classifier to produce class logits.

\end{itemize}

\begin{table*}[t]
\centering

\small
\setlength{\tabcolsep}{6pt}

\caption{Cross-Subject Classification Accuracy (\%) in Simulated Online Experiment on BNCI2014001. * indicates $p<0.05$.}
\label{tab:BNCI2014001_cross}

\begin{tabular}{L{1cm} l | ccccccccc | c}
\toprule
\textbf{Setting} & \textbf{Method} & \textbf{S0} & \textbf{S1} & \textbf{S2} & \textbf{S3} & \textbf{S4} & \textbf{S5} & \textbf{S6} & \textbf{S7} & \textbf{S8} & \textbf{Avg.} \\
\midrule
\multirow{8}{1.1cm}{\raggedright Baseline-full}
 & EEGNet        & 67.40 & 31.99 & 66.01 & 39.28 & 30.60 & 36.50 & 36.85 & 69.49 & 48.31 & $47.38_{\pm 2.51}$ \\
 & TSLANet       & 64.97 & 29.21 & 62.89 & 38.24 & 26.43 & 33.72 & 38.93 & 63.58 & 55.25 & $\mathbf{45.91}_{\pm 2.72}$ \\
 & EEGConformer  & 68.79 & 33.72 & 67.06 & 44.14 & 28.86 & 34.42 & 44.14 & 70.53 & 51.08 & $49.19_{\pm 1.33}$ \\
 & FBCNet        & 63.24 & 32.33 & 80.60 & 40.67 & 31.99 & 33.03 & 53.17 & 76.08 & 47.26 & $50.93_{\pm 1.65}$ \\
 & ATCNet        & 70.53 & 32.33 & 70.53 & 46.57 & 28.86 & 34.07 & 42.75 & 75.04 & 59.76 & $51.16_{\pm 1.44}$ \\
 & M-ShallowConvNet  & 71.22 & 33.72 & 69.49 & 39.28 & 33.03 & 32.68 & 39.28 & 73.31 & 53.17 & $49.46_{\pm 1.47}$ \\
 & mAtt          & 62.19 & 31.64 & 74.00 & 42.75 & 30.60 & 26.43 & 55.25 & 70.88 & 35.81 & $47.73_{\pm 1.79}$ \\
 & CSPNet        & 67.36 & 32.99 & 78.47 & 45.83 & 30.90 & 33.68 & 44.10 & 73.96 & 60.07 & $51.93_{\pm 1.56}$ \\
\midrule
\multirow{8}{1.4cm}{\raggedright TSE-RL}
 & EEGNet        & 65.63 & 36.81 & 66.67 & 42.71 & 32.64 & 40.97 & 39.24 & 68.06 & 58.68 & $\mathbf{50.15^{*}_{\pm 2.25}}$ \\
 & TSLANet       & 53.82 & 33.68 & 52.78 & 36.81 & 28.82 & 38.89 & 34.38 & 56.25 & 60.42 & $43.98_{\pm 2.58}$ \\
 & EEGConformer  & 59.72 & 35.42 & 62.85 & 40.97 & 30.21 & 38.54 & 45.49 & 64.24 & 65.28 & $49.19_{\pm 1.61}$ \\
 & FBCNet        & 64.24 & 36.11 & 81.94 & 43.40 & 35.42 & 34.03 & 54.86 & 77.78 & 62.85 & $\mathbf{54.51^{*}_{\pm 1.49}}$ \\
 & ATCNet        & 72.22 & 38.19 & 71.88 & 47.57 & 30.90 & 36.11 & 44.44 & 74.31 & 67.36 & $\mathbf{53.67^{*}_{\pm 1.16}}$ \\
 & M-ShallowConvNet  & 67.36 & 36.81 & 61.46 & 44.10 & 37.15 & 30.90 & 39.58 & 68.06 & 66.67 & $\mathbf{50.23}_{\pm 1.69}$ \\
 & mAtt          & 64.24 & 38.19 & 76.39 & 43.75 & 31.94 & 27.08 & 56.25 & 71.88 & 38.89 & $\mathbf{49.85^{*}_{\pm 1.21}}$ \\
 & CSPNet        & 67.71 & 36.81 & 78.47 & 47.22 & 31.94 & 35.07 & 45.14 & 73.96 & 69.10 & $\mathbf{53.96^{*}_{\pm 1.33}}$ \\
\bottomrule
\end{tabular}

\end{table*}

\textbf{Experimental settings.}
Under the \emph{Baseline-full} setting, each backbone is trained and evaluated as a purely supervised classifier on the full time windows, where no window optimization is performed. Under the \emph{TSE-RL} setting, the same pretrained backbone is split into the TSE-RL agent. The agent learns a policy to perform time-window optimization by selectively discarding or retaining windows and marking them as Onset, Sustain, or Offset. The agent then aggregates the probabilities of retained windows for classification. All experiments are independently repeated ten times, and performance is reported as mean accuracy and standard deviation across the ten runs. Statistical significance of performance improvement between TSE-RL and its corresponding Baseline-full counterpart is assessed using a paired test over the ten runs.

\textbf{Hyperparameters.}
Unless otherwise stated, we use shared hyperparameters across datasets and backbones. The history encoder $G$ operates on the most recent $k=5$ windows and is implemented as a standard Transformer encoder with one layer, four attention heads, hidden size $d_h=128$, dropout rate $0.1$. The encoded sequence is temporally mean-pooled and linearly projected to obtain a vector $h_n \in \mathbb{R}^{d_h}, d_h=64$. The FSM-constrained policy network receives the concatenated vector $z_n=[f_n;h_n;\rho_n]$ and is realized as an actor–critic multilayer perceptron with two hidden layers of 64 units. The reward and discount factor follow Section~\ref{subsec:problem_formulation}.

Training follows the three-stage scheme. In supervised pretraining, each backbone $B$ is trained end-to-end on full windows using the Adam optimizer with learning rate $10^{-3}$, batch size $64$, and $300$ epochs. After convergence, $B$ is split into $F$ and $D$, and their parameters are frozen. In policy learning, only the history encoder $G$ and the FSM-constrained policy network are optimized by PPO with learning rate $3\times10^{-4}$, PPO clipping parameter $0.2$, generalized advantage estimation parameter $\lambda=0.95$, and a total of $4\times10^{4}$ environment steps. In joint fine-tuning, $F$ are unfrozen and updated together with $G$ and the policy network using a smaller learning rate of $3\times10^{-5}$ for $2\times10^{4}$ additional environment steps.

\textbf{Configuration.}
Experiments were implemented in Python~3.10 and PyTorch~2.3.1 and ran on a workstation equipped with an NVIDIA TITAN~V GPU and an Intel Xeon Gold 6230 CPU (2.10~GHz).

\subsection{Within-subject cross-session classification in simulated online experiment}
\label{subsec:within_subject}

Tables~\ref{tab:BNCI2014001_within} and~\ref{tab:within_summary_4002_5001} report
within-subject cross-session accuracies. In all \rev{four} datasets, TSE-RL consistently
improves the performance of the Baseline-full models for most backbones, with some gains
being statistically significant. On BNCI2014001 (Table~\ref{tab:BNCI2014001_within}),
TSE-RL yields improvements for EEGNet, FBCNet, ATCNet, M-ShallowConvNet, and CSPNet, with
average accuracies increasing by about 2–3 percentage points. Only TSLANet exhibits a
slight degradation\rev{, and on this dataset it has both the lowest accuracy and the
largest standard deviation among the eight backbones. Since the rewards are
computed from the window-level confidences, such a backbone provides a noisy learning
signal for the policy}. \rev{On BNCI2014002, BNCI2015001 and OpenBMI
(Table~\ref{tab:within_summary_4002_5001}), TSE-RL improves every backbone, including
TSLANet, and the benefit persists on the much larger OpenBMI dataset.}

\subsection{Cross-subject classification in simulated online experiment}
\label{subsec:cross_subject}

Table~\ref{tab:BNCI2014001_cross} summarizes cross-subject performance on
BNCI2014001. Strong cross-subject variability makes this setting considerably
more challenging, and Baseline-full models drop substantially relative to the
within-subject case. Nevertheless, TSE-RL still improves average accuracy for
most backbones, with statistically significant gains for EEGNet, FBCNet,
ATCNet, mAtt, and CSPNet. TSE-RL is therefore compatible with a wide range of
MI backbones across both cross-session and cross-subject decoding.
\rev{Rehabilitation applications involve a further shift, as the ERD/ERS of
stroke patients differs markedly from that of young healthy subjects. On the stroke dataset of Liu et al.~\cite{liu2024stroke}, TSE-RL improves all
eight backbones (Section~5 of the Supplementary Material).}

\subsection{Visualization of learned time-window optimization}
\label{subsec:decision_viz}

\begin{figure*}[t]
    \centering
    \includegraphics[width=\textwidth]{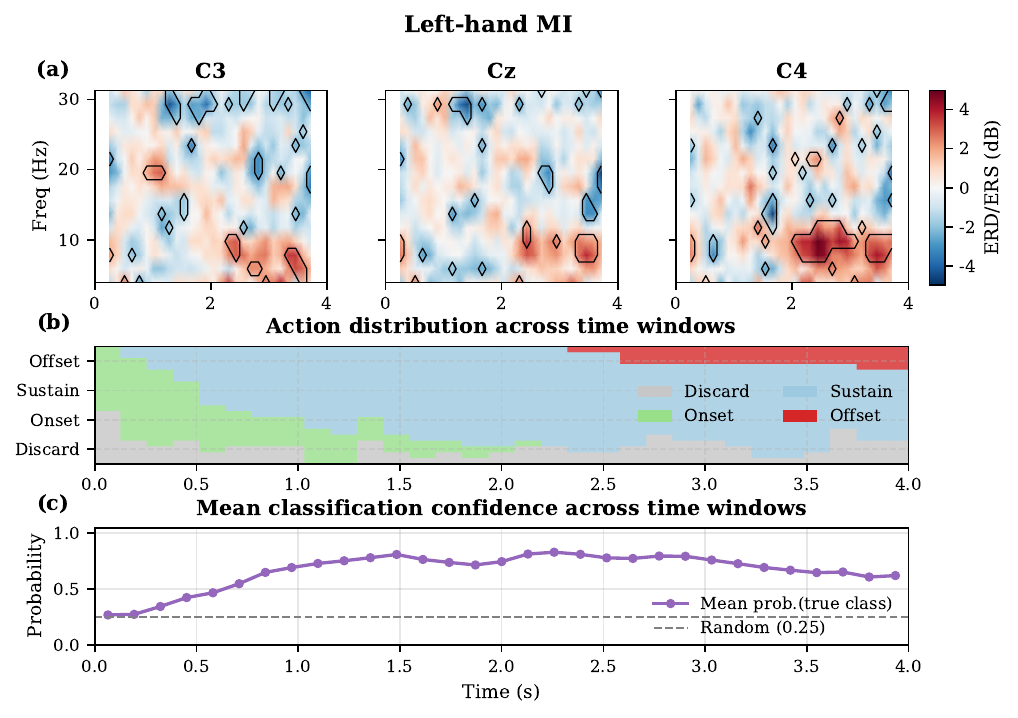}
    \caption{Average decision process for Left-hand MI on BNCI2014001. (a) Induced ERD/ERS maps computed following Graimann et al.~\cite{graimann2002visualization} with bootstrap significance testing (black contours indicate $\alpha<0.05$). (b) Action distribution over time shown as a stacked profile ordered from bottom to top as Discard, Onset, Sustain, and Offset. (c) Mean classification confidence for the true class at each time window.}
    \label{fig:avg_decision_left}
\end{figure*}

\begin{figure*}[t]
    \centering
    \includegraphics[width=\textwidth]{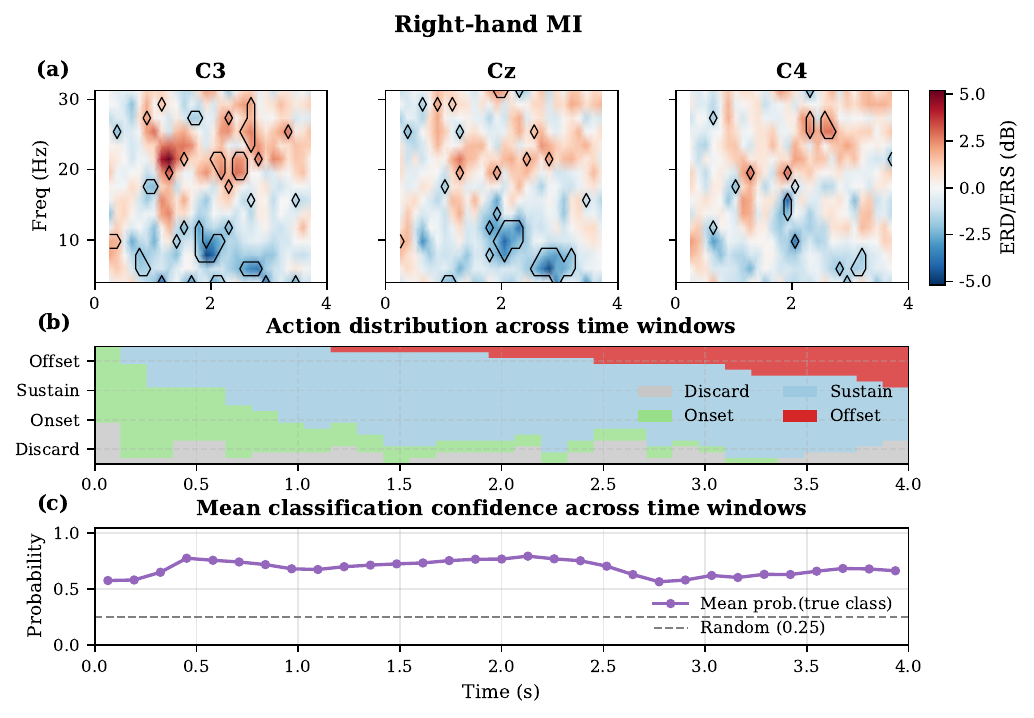}
    \caption{Average decision process for Right-hand MI of subject~0. Layout and analysis pipeline are identical to Fig.~\ref{fig:avg_decision_left}.}
    \label{fig:avg_decision_right}
\end{figure*}

The TSE-RL policy selects Onset/Sustain/Offset windows that overlap with the contralateral mu–beta ERD/ERS patterns. This suggests that it exploits the most informative segments, rather than random clipping. We visualize the learned decision process on BNCI2014001 under the simulated online setting.
A representative single-trial example is included in the \textit{Graphical Abstract} for illustration.
The analysis is then supplemented by class-wise averages to assess the generalizability of these findings beyond a single trial. Only Left-hand and Right-hand tasks are reported because these classes exhibit the clearest contralateral ERD/ERS patterns over the sensorimotor cortex. The visualizations of Feet and Tongue imagery show weaker contralateral effects with similar temporal decision patterns and are therefore omitted. All visualizations use the same configuration as in the online experiment.

Fig.~\ref{fig:avg_decision_left} and Fig.~\ref{fig:avg_decision_right} present averaged decision process for Left-hand and Right-hand motor imagery respectively. In both figures, row (a) displays induced ERD/ERS maps computed using the method from Graimann et al.~\cite{graimann2002visualization}, where phase-locked components are removed by subtracting the across-trial mean, followed by short-time Fourier transforms, baseline correction relative to the 0-1~s, and statistical validation via t-percentile bootstrap testing. Black contours highlight time-frequency regions with statistically significant deviations from baseline ($\alpha<0.05$). Row (b) visualizes the action probability distribution over time as a stacked profile, where the vertical axis represents the mean proportion of each action. The stack is ordered from bottom to top as Discard, Onset, Sustain, and Offset. Row (c) reports the mean confidence of the true class across windows, visualizing the confidence evolution throughout the trial.

The three rows reveal knowledge alignment between induced ERD/ERS patterns and the learned policy. Row (a) exhibits induced ERD/ERS over the contralateral sensorimotor cortex within the mu-beta range, with stronger desynchronization occurring during MI performing periods. In Fig.~\ref{fig:avg_decision_left}(a), Left-hand imagery exhibits a clear $\mu$-band desynchronization over C4 (around $\sim$10~Hz). In Fig.~\ref{fig:avg_decision_right}(a), Right-hand imagery shows a prominent $\beta$-band desynchronization over C3 (around $\sim$20~Hz).
Row (b) demonstrates that early time windows are dominated by Discard or Onset decisions. From approximately 0.8~s to 2.5~s, the agent's selections consist mainly of Onset and Sustain actions, indicating that discriminative evidence is concentrated in this period. Beyond approximately 2.5-4.0~s, Offset becomes the prevailing action, reducing the inclusion of low-information windows, which is consistent with typical motor imagery durations. In row (c), the mean probability of the true class rises at onset, surpasses 0.7 during the sustained desynchronization stage, and then gradually declines as neural activity returns toward baseline. These visualizations indicate that TSE-RL policy aligns its time-window optimization with task-specific induced ERD/ERS patterns from typical MI durations, by placing retained windows in periods where contralateral desynchronization is strongest, which helps explain the performance gains observed in the online experiments.

\subsection{Computational efficiency analysis}
\label{subsec:efficiency}

\begin{table}[!t]
\centering
\caption{Single-window computational efficiency under Baseline and TSE-RL settings. Time is measured in ms per window (batch size 1). FLOPs and Params are reported per forward pass.}
\label{tab:efficiency}
\small
\setlength{\tabcolsep}{4pt}
\renewcommand{\arraystretch}{1.0}
\resizebox{\columnwidth}{!}{
\begin{tabular}{@{}p{1.55cm} l c c c@{}}
\toprule
\textbf{Setting} & \textbf{Backbone} & \textbf{Time (ms)} & \textbf{FLOPs (M)} & \textbf{Params (M)} \\
\midrule
\multirow{8}{1.1cm}{\raggedright Baseline (FE+Disc)}
 & EEGNet            & 0.57 & 3.13  & 0.002 \\
 & TSLANet           & 2.73 & 7.94  & 0.366 \\
 & EEGConformer      & 4.16 & 14.39 & 0.278 \\
 & FBCNet            & 0.42 & 0.40  & 0.002 \\
 & ATCNet            & 4.00 & 12.29 & 0.036 \\
 & M-ShallowConvNet  & 0.49 & 16.37 & 0.014 \\
 & mAtt              & 5.68 & 0.29  & 0.101 \\
 & CSPNet            & 0.57 & 3.45  & 0.006 \\
\midrule
\multirow{8}{1.4cm}{\raggedright TSE-RL (Full pipeline)}
 & EEGNet            & 1.72  & 21.85 & 4.340 \\
 & TSLANet           & 3.77  & 42.77 & 5.400 \\
 & EEGConformer      & 9.24  & 99.32 & 1.940 \\
 & FBCNet            & 1.75  & 5.00  & 0.337 \\
 & ATCNet            & 5.55  & 85.28 & 0.199 \\
 & M-ShallowConvNet  & 2.42  & 99.36 & 0.198 \\
 & mAtt              & 16.67 & 8.25  & 4.979 \\
 & CSPNet            & 1.74  & 24.33 & 6.660 \\
\bottomrule
\end{tabular}
}
\end{table}

\begin{table*}[t]
\centering
\small
\setlength{\tabcolsep}{6pt}
\renewcommand{\arraystretch}{1.0}

\caption{Ablation study on the EEGNet backbone  across settings and MI datasets. Mean\,\,$\pm$\,s.d. classification accuracy (\%).}
\label{tab:ablation}

\begin{tabular}{@{}ccccccc@{}}
\toprule
\multicolumn{3}{c}{\textbf{Components}}
  & \multicolumn{2}{c}{\textbf{BNCI2014001}}
  & \multicolumn{1}{c}{\textbf{BNCI2014002}}
  & \multicolumn{1}{c}{\textbf{BNCI2015001}} \\
\cmidrule(r){1-3}\cmidrule(lr){4-5}\cmidrule(lr){6-6}\cmidrule(l){7-7}
\textbf{RL} & \textbf{Action mask} & \textbf{History enc.}
  & \textbf{within-subject}
  & \textbf{cross-subject}
  & \textbf{within-subject}
  & \textbf{within-subject} \\
\midrule
$\times$ & $\times$ & $\times$
  & $70.15_{\pm 2.04}$
  & $47.38_{\pm 2.51}$
  & $85.59_{\pm 1.91}$
  & $82.42_{\pm 1.96}$ \\
$\checkmark$ & $\times$ & $\times$
  & $69.10_{\pm 2.11}$
  & $48.07_{\pm 2.58}$
  & $86.54_{\pm 1.81}$
  & $83.32_{\pm 1.80}$ \\
$\checkmark$ & $\checkmark$ & $\times$
  & $70.75_{\pm 1.82}$
  & $48.72_{\pm 2.36}$
  & $86.86_{\pm 1.78}$
  & $83.87_{\pm 1.71}$ \\
$\checkmark$ & $\times$ & $\checkmark$
  & $71.41_{\pm 1.86}$
  & $49.48_{\pm 2.17}$
  & $87.02_{\pm 1.81}$
  & $84.25_{\pm 1.87}$ \\
$\checkmark$ & $\checkmark$ & $\checkmark$
  & $\mathbf{72.34_{\pm 1.91}}$
  & $\mathbf{50.15_{\pm 2.25}}$
  & $\mathbf{87.62_{\pm 1.67}}$
  & $\mathbf{84.88_{\pm 1.71}}$ \\
\bottomrule
\end{tabular}

\end{table*}

Online MI decoding requires repeated inference on sliding windows, so the computational cost for each window is critical for practical deployment. We compare the computational efficiency of different backbones. A single EEG window of size $22 \times 250$ (1.0~s at 250~Hz) from BNCI2014001 is forwarded with batch size 1.
Table~\ref{tab:efficiency} reports inference time, forward FLOPs, and parameter count. \textit{Baseline} measures the backbone classifier only, consisting of a feature extractor and a discriminator (FE+Disc). \textit{TSE-RL} measures the full window optimization pipeline, which adds a history encoder and a policy network on top of FE+Disc.
The increase in FLOPs and parameter count from Baseline to TSE-RL primarily results from the additional history encoder and policy network. They process multiple historical windows with intermediate high-dimensional features. Despite this additional computation, the per-window inference time remains low across all backbones. Note that mAtt exhibits longer inference time due to its reliance on eigendecomposition operations, which are less suited to parallel computations on GPUs. Even the slowest configuration remains well below the 0.1~s stride interval and significantly below the 1.0~s window duration. This indicates that TSE-RL introduces only a modest overhead and is compatible with online decoding requirements.

\subsection{Ablation study}
\label{subsec:ablation}

To validate the core contributions of TSE-RL, an ablation study is conducted on the EEGNet backbone. The results are reported in Table~\ref{tab:ablation} across settings and datasets. The comparison starts from a supervised baseline that disables RL, the history encoder, and the FSM-based action mask, which corresponds to EEGNet (Baseline-full) operating on all windows without window optimization. RL is then enabled while keeping both the history encoder and the action mask disabled, yielding a minimal policy that uses only the current-window feature and the trial-progress indicator. Two intermediate variants are further evaluated by adding either the FSM-based action mask or the history encoder. And the full model then enables RL, the history encoder, and the action mask simultaneously. Overall, Table~\ref{tab:ablation} shows that RL alone improves performance over the supervised baseline, indicating the benefit of learning an time-window optimization policy. Adding either the FSM-based action mask or the history encoder provides further gains, and enabling both consistently achieves the highest accuracy across all datasets and settings.

\begin{figure}[!tb]
    \centering
    \includegraphics[width=\columnwidth]{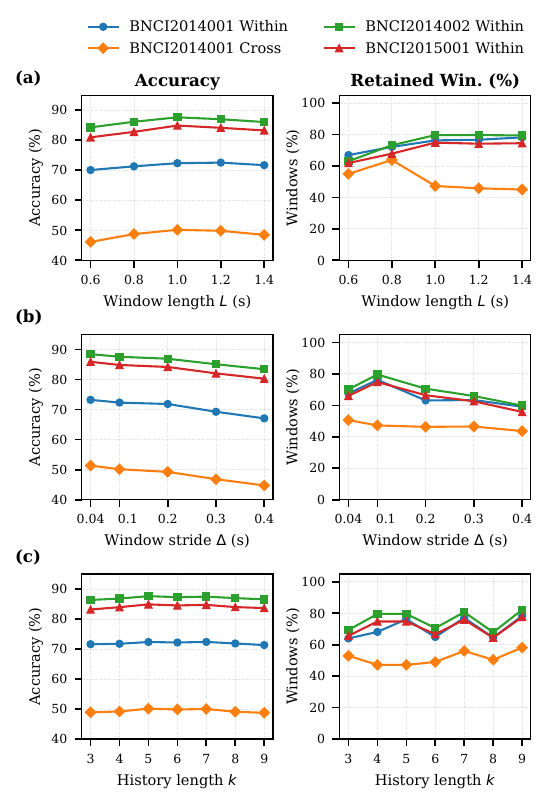}
    \caption{Sensitivity analysis of window length $L$, stride step $\Delta$, and history length $k$ on BNCI2014001 (within-subject), BNCI2014001 (cross-subject), BNCI2014002 (within-subject), and BNCI2015001 (within-subject). The left column displays classification accuracy, while the right column shows the ratio of retained windows.}
    \label{fig:sensitivity}
\end{figure}

\subsection{Sensitivity analysis}
\label{subsec:sensitivity}

To assess the robustness of TSE-RL to hyperparameters, a sensitivity analysis is performed with respect to the window length $L$, stride $\Delta$, and history length $k$ on the EEGNet backbone. The default configuration in the main experiments is $L=1.0$~s, $\Delta=0.1$~s, and $k=5$. Fig.~\ref{fig:sensitivity} varies these hyperparameters within practical ranges ($L\in[0.6,1.4]$~s, $\Delta\in[0.04,0.4]$~s, and $k\in[3,9]$) and reports accuracy along with the retained-window ratio (retained windows divided by all candidate windows). This ratio is reported because changing $L$ or $\Delta$ alters the number of candidate windows per trial, making raw window counts incomparable. Overall, Fig.~\ref{fig:sensitivity} reveals that TSE-RL demonstrates considerable stability around the default parameter configuration, with only moderate performance variation under perturbations of $L$, $\Delta$, and $k$.

\section{Conclusion}
\label{sec:conclusion}

Online MI-BCI decoding operates on continuous EEG dataflow using fixed-length
sliding windows, some of which are misaligned with the true boundaries of MI
events. Time window optimization therefore matters for forming an informative
segment by adaptively accumulating relevant windows and discarding irrelevant
ones. However, most existing strategies evaluate candidate windows
independently, ignoring temporal dependencies across consecutive windows and
yielding fragmented or physiologically implausible sequences.
To address this limitation, we proposed Temporal Structure Embedded
Reinforcement Learning, which reformulates window optimization as a
constrained Markov Decision Process in which an FSM-constrained action mask
keeps the selected windows consistent with the temporal structure of MI events
while assigning each of them an explicit temporal role. A Transformer-based
history encoder captures long-range inter-window dependencies with better
training stability than recurrent designs. The framework is plug-and-play and
integrates with diverse MI classification backbones. Extensive experiments show
consistent improvements in cross-session and cross-subject settings at modest
overhead compatible with real-time constraints. Visualization analyses further
show that the learned policy aligns decision windows with contralateral
mu-beta ERD/ERS, confirming that it exploits neurophysiological patterns.
These results establish TSE-RL as an effective approach for online time window
optimization. \rev{Its behavior under closed-loop feedback remains to be
examined, since the simulated online experiment reproduces the non-stationary
environmental noise and the fatigue-induced neural pattern shifts of real
recordings but leaves the BCI output disconnected from the user. Closed-loop
validation on data acquired will be studied in future
work.}

\section*{References}
\bibliographystyle{IEEEtran}
\bibliography{Reference}

\end{document}